\documentclass[letterpaper]{article} 
\usepackage{aaai25}  
\usepackage{times}  
\usepackage{helvet}  
\usepackage{courier}  
\usepackage[hyphens]{url}  
\usepackage{graphicx} 
\usepackage{natbib}  
\usepackage{caption} 
\usepackage{algorithm}
\usepackage{algorithmic}
\usepackage{multirow}
\usepackage{xcolor}
\usepackage{cite}
\usepackage{amsmath}
\usepackage{amsthm}
\usepackage{amssymb}
\usepackage{graphicx}
\usepackage{multirow}
\usepackage{booktabs}
\usepackage{bbding}

\usepackage{newfloat}
\usepackage{listings}
\DeclareCaptionStyle{ruled}{labelfont=normalfont,labelsep=colon,strut=off} 
\floatstyle{ruled}
\newfloat{listing}{tb}{lst}{}
\floatname{listing}{Listing}
\title{MLAAN: Scaling Supervised Local Learning \\with Multilaminar Leap Augmented Auxiliary Network}
\author{
    Yuming Zhang\textsuperscript{\rm 1}\equalcontrib,
    Shouxin Zhang\textsuperscript{\rm 1}\equalcontrib, 
    Peizhe Wang\textsuperscript{\rm 1}\equalcontrib, 
    Feiyu Zhu\textsuperscript{\rm 2}, 
    Dongzhi Guan\textsuperscript{\rm 1\textdagger},
    Junhao Su\textsuperscript{\rm 1}, 
    Jiabin Liu\textsuperscript{\rm 1}, 
    Changpeng Cai\textsuperscript{\rm 1}\thanks{Corresponding Author: Dongzhi Guan and Changpeng Cai.}
}
\affiliations{
    \textsuperscript{\rm 1}Southeast University\\
    Nanjing 211189, China\\
    \{220221309, 220241560, 220221387, guandongzhi, 220211548, 101005394, 220221408\}@seu.edu.cn\\
    \textsuperscript{\rm 2}University of Shanghai for Science and Technology\\
    Shanghai 200093, China\\
    zeurdfish@gmail.com

}

\usepackage{bibentry}

\begin{document}

\maketitle

\begin{abstract}
Deep neural networks (DNNs) typically employ an end-to-end (E2E) training paradigm which presents several challenges, including high GPU memory consumption, inefficiency, and difficulties in model parallelization during training. Recent research has sought to address these issues, with one promising approach being local learning. This method involves partitioning the backbone network into gradient-isolated modules and manually designing auxiliary networks to train these local modules. Existing methods often neglect the interaction of information between local modules, leading to myopic issues and a performance gap compared to E2E training. To address these limitations, we propose the Multilaminar Leap Augmented Auxiliary Network (MLAAN). Specifically, MLAAN comprises Multilaminar Local Modules (MLM) and Leap Augmented Modules (LAM). MLM captures both local and global features through independent and cascaded auxiliary networks, alleviating performance issues caused by insufficient global features. However, overly simplistic auxiliary networks can impede MLM's ability to capture global information. To address this, we further design LAM, an enhanced auxiliary network that uses the Exponential Moving Average (EMA) method to facilitate information exchange between local modules, thereby mitigating the shortsightedness resulting from inadequate interaction. The synergy between MLM and LAM has demonstrated excellent performance. Our experiments on the CIFAR-10, STL-10, SVHN, and ImageNet datasets show that MLAAN can be seamlessly integrated into existing local learning frameworks, significantly enhancing their performance and even surpassing end-to-end (E2E) training methods, while also reducing GPU memory consumption. 
\begin{links}
  \link{Code}{https://github.com/Wendy231015/MLAAN}
\end{links}
\end{abstract}

\begin{figure}[h]
\centering
\includegraphics[width=0.91\columnwidth]{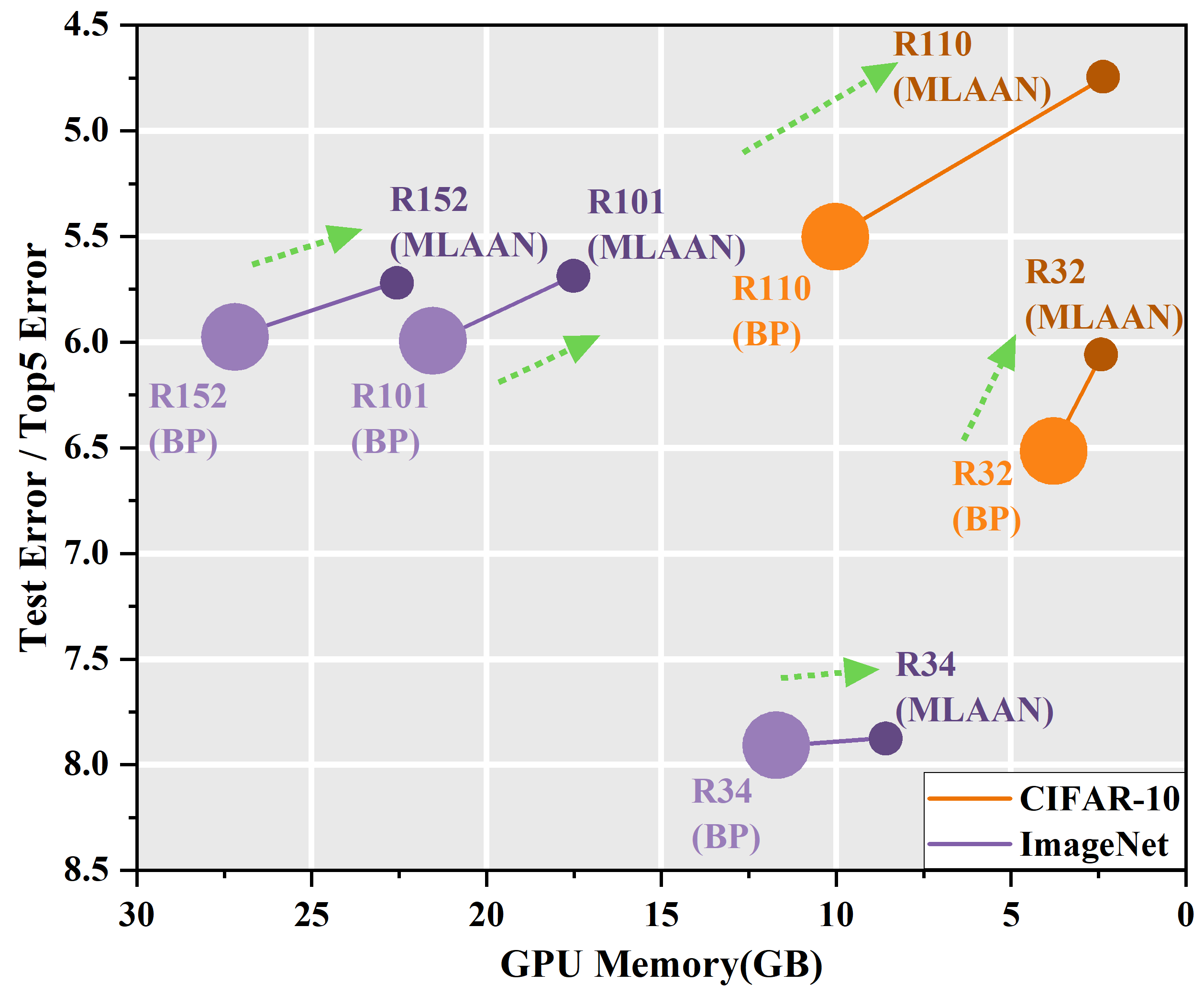} 
\caption{Comparison between different methods with MLAAN and BP regarding GPU Memory, Test Error, and Top5 Error. Results are obtained using ResNet-32 and ResNet-110 on CIFAR-10, ResNet-34, ResNet-101, and ResNet-152 on ImageNet.}
\label{fig:1}
\end{figure}

\section{Introduction}
End-to-end (E2E) backpropagation, a mainstream method for training deep neural networks, has achieved remarkable success in image segmentation \cite{r14}, target tracking \cite{r16}, visual sound separation \cite{r18}, and other fields. Despite these accomplishments, E2E learning has its drawbacks. It necessitates retaining the complete computational graph and all intermediate activations post each local module's computation \cite{r19}, increasing GPU memory usage. Additionally, the E2E structure delays parameter updates in initial layers until the entire forward and backward propagation cycles conclude, causing inefficiency \cite{r21}, \cite{r22}. This sequential gradient processing also impedes model parallelization \cite{r23}, \cite{r24} and raises biological plausibility concerns due to its reliance on cross-layer backpropagation \cite{r25}, \cite{r26}.

Due to the inherent flaws of E2E training \cite{r27}, \cite{r29}, some researchers are exploring local learning as an alternative. Local learning divides the network into gradient-isolated local modules and designs auxiliary networks to assist their training \cite{r30}. This method eliminates the need for sharing gradients and activation parameters between modules, reducing GPU memory requirements \cite{r19}. In addition, local learning uses local error signals, which align more closely with biological feasibility principles. However, the existing focus on local features and the lack of information exchange between local modules significantly lower overall performance than that of E2E training \cite{r31}. Despite some recent improvements in the performance of local learning \cite{r24}, \cite{r28}, current methods still have myopia issues and insufficient global feature representation, limiting their application in precision-critical tasks. Therefore, the crucial goal of local learning is to ensure that each local module receives ample global information and facilitates information exchange between modules. Maintaining GPU memory efficiency while improving global performance is essential for making local learning a viable alternative to E2E training.

\begin{figure*}[!t]
\begin{minipage}[b]{1.0\linewidth}
  \centering
  \includegraphics[width=16.0cm]{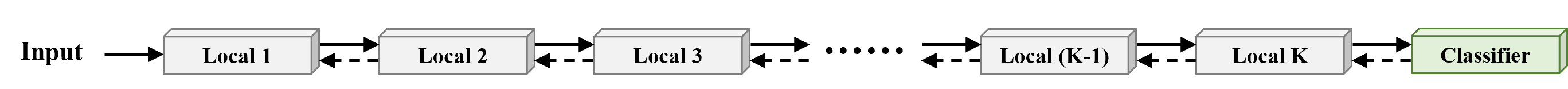}
  \centerline{(a)}\medskip
\end{minipage}
\begin{minipage}[b]{1.0\linewidth}
  \centering
  \includegraphics[width=16cm]{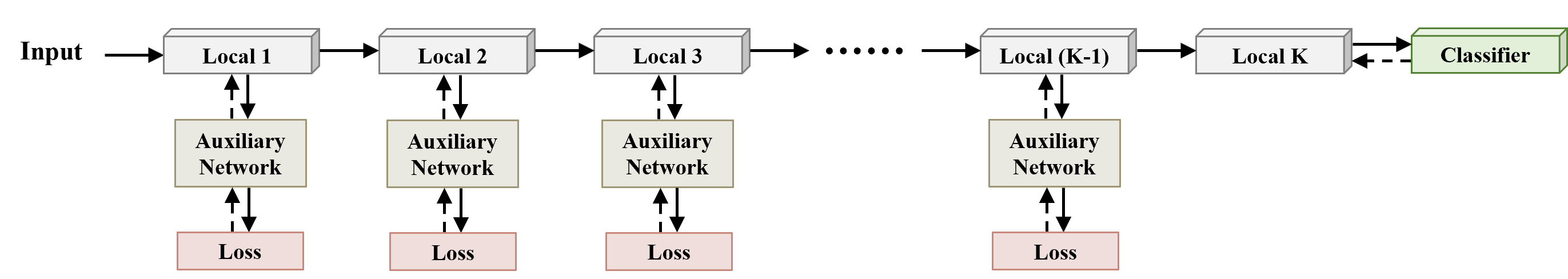}
  \centerline{(b)}\medskip
\end{minipage}
\begin{minipage}[b]{1.0\linewidth}
  \centering
  \includegraphics[width=16.0cm]{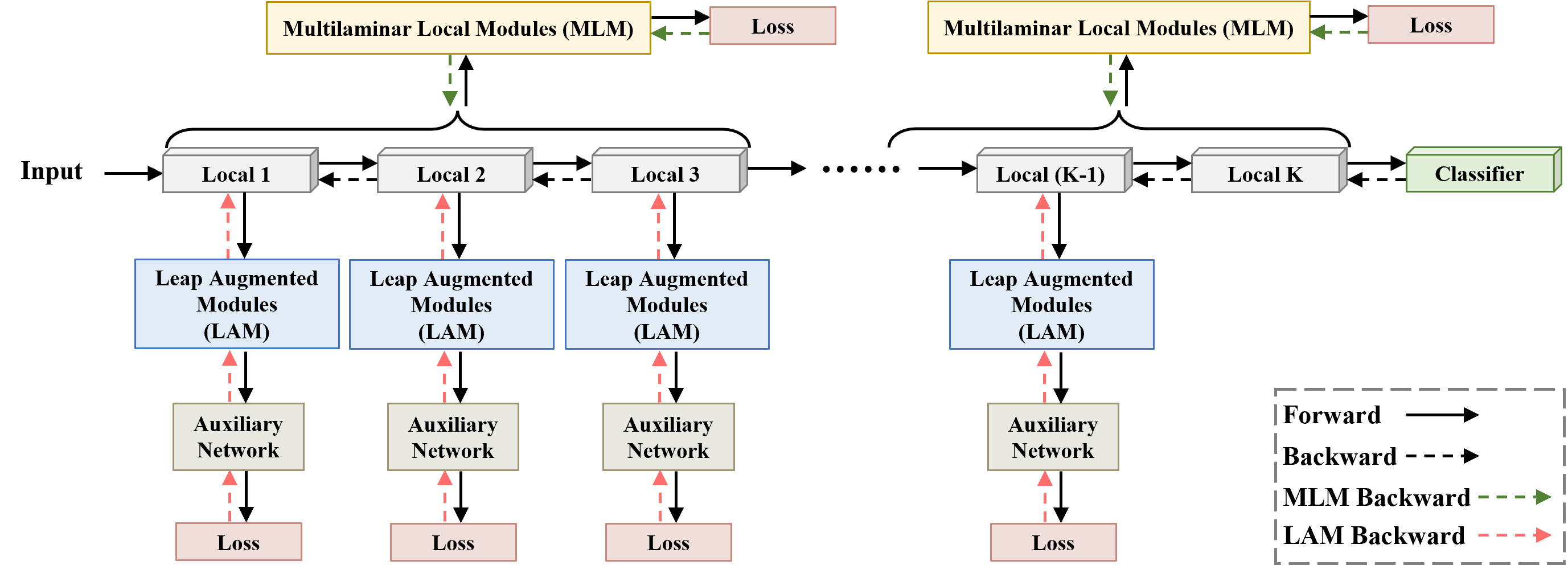}
  \centerline{(c)}\medskip
\end{minipage}
\caption{Comparison of (a) E2E backpropagation, (b) other supervised local learning methods, and (c) our proposed method. The details of our method are in Figure \ref{fig:3}.}
\label{fig:2}
\end{figure*}

In this paper, we propose the Multilaminar Leap Augmented Auxiliary Network (MLAAN), a novel supervised local learning method, as illustrated in Figure \ref{fig:2}. MLAAN employs a dual architecture where the Multilaminar Local Modules (MLM) design divides the network into independent and cascaded levels. The independent level uses standalone auxiliary networks to capture local features, while the cascaded level utilizes cascade auxiliary networks to gain global features through module inter-sharing. The two levels complement each other and sufficiently learn global and local features. Furthermore, we observe that even with multiple local modules sharing weights, the backbone network can still be misled by overly simplistic auxiliary networks, leading to shortsightedness. To resolve this issue, we further design the Leap Augmented Modules (LAM). LAM tackles the problem of weakened supervision caused by overly simplistic auxiliary networks. This module utilizes the network structure of subsequent layers and updates using the Exponential Moving Average (EMA). LAM is strategically placed between the primary and secondary networks within each gradient-isolated module to mitigate myopia. The synergy between MLM and LAM significantly enhances network performance. The efficacy of MLAAN is validated on a suite of benchmark image classification datasets, including CIFAR-10 \cite{r1}, STL-10 \cite{r3}, SVHN \cite{r2}, and ImageNet \cite{r4}. Experimental results demonstrate that MLAAN almost perfectly resolves the existing issues in local learning.

\begin{itemize}
\item We propose a novel framework, MLAAN, which effectively overcomes the traditional limitations of local learning, particularly the challenge of accessing adequate global information and the myopic perspective resulting from limited information exchange.
\item MLAAN is a versatile plug-and-play module that can be seamlessly integrated into existing supervised local learning methods, significantly enhancing their performance.
\item MLAAN's effectiveness has been validated on widely used datasets, including CIFAR-10, STL-10, SVHN, and ImageNet, achieving state-of-the-art (SOTA) performance. Remarkably, it surpasses the performance of E2E training while also substantially saving GPU memory.
\end{itemize}

\section{Related Works}
\subsection{Alternatives of Backpropagation}

Traditional E2E learning faces challenges like high GPU memory consumption, reduced training efficiency, difficulties in model parallelization, and poor biological plausibility. Researchers are developing alternative strategies, such as target propagation \cite{r35,r36,r37}, feedback alignment \cite{r38,r39}, synthetic gradients \cite{r40}, feature replay \cite{r41}, and decoupled parallel backpropagation \cite{r42}, to tackle memory and efficiency issues. Some suggest using forward gradient learning instead of backpropagation \cite{r43}. However, most of these methods still rely on global loss values and have not effectively addressed challenges with large-scale datasets like ImageNet \cite{r4}.

\subsection{Local Learning}

Local learning is considered a promising alternative to traditional E2E methods \cite{r28, r50}. However, the lack of local inter-module interaction and poor global performance has hindered its development \cite{r30, r53}. To improve global performance, some scholars have used self-supervised loss to preserve features during local module propagation. For example, Y. Xiong et al. \cite{r47} designed SimCLR loss for Local Contrastive Learning. Sindy et al. \cite{r48} applied Comparative Predictive Coding loss to reduce feature loss in Greedy InfoMax. These methods perform well with a few network blocks but struggle with many blocks. Adrien et al.'s method \cite{r49}, combining Hebbian learning with local learning, addresses this issue but faces challenges on large datasets. Methods like InfoPro \cite{r19} and DGL \cite{r24} have improved local learning performance on large datasets such as ImageNet \cite{r4}. However, due to gradient isolation, their performance still falls short of E2E methods \cite{r51,r52}. Enhancing inter-module communication to match or surpass E2E performance remains a critical issue. 

Based on existing research and shortcomings, this paper aims to design a new method to improve inter-module communication and address model myopia, ultimately enhancing supervised local learning performance and enabling advanced training of large models on resource-limited platforms.

\section{Method}
\subsection{Preparations}

To begin, we present an overview of the conventional E2E BP-supervised learning paradigm and the BP mechanism for contextual clarity. Let $x$ and $y$ denote a data sample and its corresponding true label, respectively. The neural network with parameters $\theta$ is represented as $f_{\theta}$, with its forward computation denoted as $f(\cdot)$. In BP, we evaluate the loss function ${\mathcal{L}}(\hat{y}, y)$ between the output of the last module and the ground truth label, and propagate it back iteratively to the preceding blocks.

Supervised local learning divided the network into multiple local modules, the output of the $j$-th local module is used as the input of the subsequent $(j + 1)$-th local module. During forward propagation, $x_{j+1}=f_{\theta_{j}}\left(x_{j}\right)$. 
For existing supervised local learning methods, the output of a local module is directed to its auxiliary network to generate local monitoring signals $\hat{y}_{j}=g_{\gamma_{j}}\left(f_{\theta_{j}}\left(x_{j}\right)\right)$, where $g_{\gamma_{j}}(\cdot)$ represents the auxiliary network updating function. The updates are made based on the following equations:
\begin{equation}
\gamma_{j} \leftarrow \gamma_{j}-\eta_{\alpha} \times \nabla_{\gamma_{j}} \mathcal{L}\left(\hat{y}_{j}, y\right)
\end{equation}
\begin{equation}
\theta_{j} \leftarrow \theta_{j}-\eta_{\iota} \times \nabla_{\theta_{j}} \mathcal{L}\left(\hat{y}_{j}, y\right)
\end{equation}
where $\gamma_{j}, \theta_{j}$ means updated parameters of the auxiliary networks and the local modules, $\eta_{\alpha}$ and $\eta_{\iota}$ denote the learning rate of them. 

\subsection{Multilaminar Local Modules}
\begin{figure*}[!t]
\centering
\includegraphics[width=1.61\columnwidth]{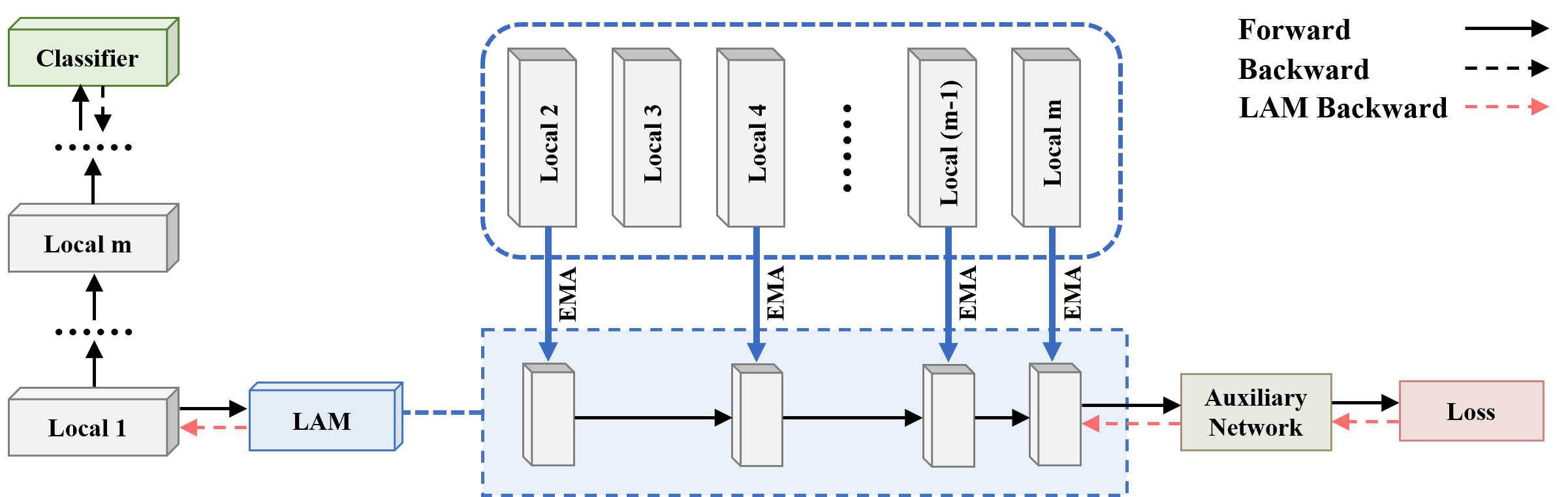}%
\caption{The Leap Augmented Modules architecture. As the proximity to the early blocks increases, the utilization of auxiliary layers employing EMA decreases.}
\label{fig:3}
\end{figure*}
Current supervised local learning methods face certain limitations in that they rely on auxiliary networks to independently update each gradient-isolated local module, which restricts the local modules from acquiring global features. Therefore, we propose Multilaminar Local Modules (MLM), designed to facilitate local modules in capturing global features while retaining local features as much as possible.

In our MLM, we divide the entire network into two levels of local modules: cascaded modules and independent modules. Each local module functions independently, while multiple adjacent local modules form a cascaded module. 

We assume that each cascaded module computes $K(K>1)$ consecutive local modules. For the $j$-th  local module within a cascade module, denoted as $f_{\theta_{j}}$, it receives signals from its independent auxiliary network $g_{\gamma_{j}}$ and $k$ cascaded auxiliary networks, $h_{\beta_{i}}, \cdots, h_{\beta_{i+k-1}}$. The local supervision of a particular local module occurs $(K+1)$  times, once from the loss function of independent modules, $\mathcal{L}\left(\hat{y}_{j}, y\right)$, and k times from loss function of cascaded modules, $\mathcal{L}\left(\hat{y}_{i}, y\right), \cdots, \mathcal{L}\left(\hat{y}_{i+k-1}, y\right)$. 

The updated rules can be summarized as follows:
\begin{equation}
\gamma_{j} \leftarrow \gamma_{j}-\eta_{\alpha} \times \nabla_{\gamma_{j}} \mathcal{L}\left(\hat{y}_{j}, y\right)
\end{equation}
\begin{equation}
\beta_{j} \leftarrow \beta_{j}-\sum_{n=i}^{i+k-1}\left(\eta_{c} \times \nabla_{\beta_{j}} \mathcal{L}\left(\hat{y}_{n}, y\right)\right)
\end{equation}
\begin{equation}
\theta_{j} \leftarrow \theta_{j}-\eta_{\iota} \times \nabla_{\theta_{j}} \mathcal{L}\left(\hat{y}_{j}, y\right)
\end{equation}
where $\theta_{j}$ means updated parameters of the local modules, $\gamma_{j}$, $\beta_{j}$ denote updated parameters of the independent and cascaded auxiliary networks, $\eta_{l}$, $\eta_{\alpha}$, and $\eta_{c}$ are the learning rates of them, respectively. 

\subsection{Leap Augmented Modules}

After applying MLM, we observe that even when multiple local modules share weights, the backbone network can still be misled by overly simplistic auxiliary networks, leading to insufficient supervision of the hidden layers by local modules and suboptimal learning outcomes. To handle this concern, we introduce generic Leap Augmented Modules(LAM). LAM is designed to enhance the exchange of information between the current local module and subsequent local modules, thereby promoting improved learning outcomes.

As shown in Figure \ref{fig:3}, for the $j$-th local module $f_{\theta_{j}}$, its auxiliary network is denoted as $g_{\gamma_{j}}$. We assume that $f_{\theta_{j}}$ is expected to receive information from $p(p \geq 1)$ hidden layers, distributed across the early, intermediate, and deeper layers of the subsequent local modules. To illustrate how our method is implemented, we assume $p=1$, meaning that the auxiliary network selects only one hidden layer from subsequent local modules to serve as the enhanced auxiliary network. We perform a deepcopy operation: $\gamma_{j}^1 =\operatorname{deepcopy}\left(\theta ^{j+i}\right)$$(i \geq 1 )$, where $\theta ^{j+i}$ represents the hidden layer extracted in the $(j+i)$-th local module. The $\gamma_{j}^1$, receives the information from the hidden layer of the $(j+i)$-th local modules and is updated using Exponential Moving Average (EMA). The updated rules are as follows:
\begin{equation}
\gamma_j^1 \leftarrow \gamma_j^1 - \eta_\alpha \times \nabla_{\gamma_j^1} \mathcal{L}(\hat{y_j}, y)
\end{equation}
\begin{equation}
\gamma_j^1 \leftarrow EMA(\gamma_j^1, \theta_{j+i})
\end{equation}
\begin{equation}
\theta_j \leftarrow \theta_j - \eta_l \times \nabla_{\theta_j} \mathcal{L}(\hat{y_j}, y)
\end{equation}
where $\gamma_j^1$ represents the parameters of the first layer of the $j$-th auxiliary network and $\theta_j$ represents those of the $j$-th local module, $\eta_{\alpha}$ and $\eta_{l}$ are the learning rates of them. After updating with local gradients, $\gamma_j^1$ undergoes further refinement by incorporating the parameters of the $(j+i)$-th local module $\theta_{j+i}$ via the EMA, which is a weighted sum operation.

Moreover, when $p=2$, deeper integration is employed by performing a deepcopy operation: $\gamma_{j}^2 =\operatorname{deepcopy}\left(\theta ^{j+i+k}\right)$$(i,k \geq 1 )$, this operation utilizes the information from the $(i+j+k)$-th local module, allowing for even more robust incorporation of global information into the auxiliary network. When $p$ is larger, the construction of the auxiliary network follows this pattern accordingly. And the $j$-th auxiliary network will be represented as $\gamma_j=\{\gamma_j^1,\gamma_j^2,\cdots\}$ finally.

\subsection{Multilaminar Leap Augmented Auxiliary Network}
\begin{table*}[!t]
\centering
\setlength{\tabcolsep}{0.8mm}
\begin{tabular}{cccccc}
\toprule
\multirow{2}{*}{Dataset} &
  \multirow{2}{*}{Method} &
  \multicolumn{2}{c}{ResNet-32} &
  \multicolumn{2}{c}{ResNet-110} \\ \cmidrule{3-4} \cmidrule{5-6}
 &                   & K=8(Test Error) & K=16(Test Error)               & K=32(Test Error) & K=55(Test Error)                \\ \midrule
\multirow{6}{*}{CIFAR-10} &
  DGL &
  11.63±0.21 &
  14.08±0.31 &
  12.51±0.19 &
  14.45±0.54 \\
 & \textbf{DGL*}     & \textbf{6.25±0.12(↓5.26-5.50)}       & \textbf{6.36±0.15(↓7.57-7.87)} & \textbf{5.38±0.27(↓6.86-7.40)}        & \textbf{5.42±0.52(↓8.51-9.03)}  \\
 & InfoPro           & 11.51±0.33                & 12.93±0.38                          & 12.26±0.28                 & 13.22±0.42                           \\
 & \textbf{InfoPro*} & \textbf{6.37±0.50(↓4.64-5.14)}       & \textbf{6.37±0.35(↓6.21-6.91)} & \textbf{5.41±0.49(↓6.36-7.34)}        & \textbf{5.60±0.55(↓7.07-7.62)}  \\
 & \textbf{MLAAN}    & \textbf{6.36±0.23}       & \textbf{6.51±0.55}             & \textbf{5.41±0.62}        & \textbf{6.02±0.67}              \\ \cmidrule{2-6}
 & \multicolumn{5}{c}{\centering BP(ResNet-32)=6.37±0.43, BP(ResNet-110)=5.42±0.31} \\ \midrule
\multirow{6}{*}{STL-10} &
  DGL &
  25.05±0.20 &
  27.14±0.19 &
  25.67±0.26 &
  28.16±0.31 \\
 & \textbf{DGL*}     & \textbf{19.35±0.51(↓5.19-6.21)}       & \textbf{19.02±0.11(↓8.01-8.23)}          & \textbf{19.54±0.34(↓5.79-6.47)}        & \textbf{19.66±0.55(↓7.95-8.50)} \\
 & InfoPro           & 27.32±0.22                & 29.28±0.15                          & 28.58±0.20                 & 29.20±0.17                           \\
 & \textbf{InfoPro*} & \textbf{19.21±0.19(↓7.92-8.30)}       & \textbf{19.21±0.12(↓9.95-10.19)}          & \textbf{19.51±0.16(↓8.91-9.23)}        & \textbf{19.24±0.21(↓9.75-10.17)}           \\
 & \textbf{MLAAN}    & \textbf{19.34±0.15}       & \textbf{19.96±0.53}            & \textbf{20.46±0.64}        & \textbf{19.66±0.67}             \\ \cmidrule{2-6}
 & \multicolumn{5}{c}{\centering BP(ResNet-32)=19.35±0.24, BP(ResNet-110)=19.67±0.34} \\ \midrule
\multirow{6}{*}{SVHN} &
  DGL &
  4.83±0.11 &
  5.05±0.08 &
  5.12±0.16 &
  5.36±0.19 \\
 & \textbf{DGL*}     & \textbf{2.79±0.12(↓1.92-2.16)}       & \textbf{2.73±0.19(↓2.13-2.51)}           & \textbf{2.78±0.22(↓2.12-2.56)}        & \textbf{2.90±0.14(↓2.32-2.46)}  \\
 & InfoPro           & 5.61±0.27                & 5.97±0.17                           & 5.89±0.19                 & 6.11±0.22                            \\
 & \textbf{InfoPro*} & \textbf{2.97±0.18(↓2.46-2.82)}       & \textbf{2.95±0.2(↓2.82-3.02)}  & \textbf{2.87±0.32(↓2.70-3.34)}        & \textbf{2.90±0.42(↓2.79-3.21)}  \\
 & \textbf{MLAAN}    & \textbf{2.85±0.22}       & \textbf{2.82±0.38}             & \textbf{2.79±0.31}        & \textbf{2.89±0.31}              \\ \cmidrule{2-6} 
 & \multicolumn{5}{c}{\centering BP(ResNet-32)=2.99±0.18, BP(ResNet-110)=2.92±0.22} \\ \bottomrule
\end{tabular}
\caption{Comparison of supervised local learning methods and BP on image classification datasets. The averaged test errors are reported from three independent trials. The * means the addition of our MLAAN.}
\label{tab:table1}
\end{table*}
MLM and LAM each possess distinct characteristics. MLM captures both local and global features through independent and cascaded auxiliary networks. LAM is further designed to accurately capture local features while avoiding misleading the main network caused by overly simplistic auxiliary networks. Based on their characteristics, we fuse the two approaches to make it a Multilaminar Leap Augmented Auxiliary Network (MLAAN). MLAAN performs exceptionally well under the synergy between MLM and LAM.

One local module, denoted as $f_{\theta_{s}}$, within its cascaded module, $h_{\beta_{s}}=\{h_{\beta_{s}}, h_{\beta_{s+1}}, \cdots, h_{\beta_{s+k-1}}\}$. With the application of MLM, $f_{\theta_{s}}$ receives $(K+1)$ local supervisions. Upon incorporating the LAM, the parameters of $p$ hidden layers extracted in corresponding local modules, $\theta_{s+P}=\{\theta_{s+1},\theta_{s+2},\cdots,\theta_{s+p}\}$, are integrated into the $s$-th auxiliary network, $\gamma_s=\left\{\gamma_{s}^{1},\gamma_{s}^{2},\cdots,\gamma_{s}^{p}\right\}$. MLM and LAM work in parallel to maximize the utilization of global and local features. 

The overall updated rules of MLAAN are described as follows:
\begin{equation}
\gamma_s \leftarrow \gamma_s - \eta_\alpha \times \nabla_{\gamma_s} \mathcal{L}(\hat{y_j}, y)
\end{equation}
\begin{equation}
\gamma_s \leftarrow EMA(\gamma_s, \theta_{s+M})
\end{equation}
\begin{equation}
\beta_{s} \leftarrow \beta_{s}-\sum_{n=i}^{i+k-1}\left(\eta_{c} \times \nabla_{\beta_{s}} \mathcal{L}\left(\hat{y}_{n}, y\right)\right)
\end{equation}
\begin{equation}
\theta_{s} \leftarrow \theta_{s}-\eta_{\iota} \times \nabla_{\theta_{s}} \mathcal{L}\left(\hat{y}_{s}, y\right)
\end{equation}
where $\theta_{s}$ means the parameters of the $s$-th local module, $\gamma_{s}$ and $\beta_{s}$ denote the parameters of its independent and cascaded auxiliary networks, $\eta_{l}$, $\eta_{\alpha}$, and $\eta_{c}$ are the learning rates of them, respectively. 

\section{Experiments}
\subsection{Experimental Setup}

We conduct experiments with ResNet \cite{r5} architectures of various depths on CIFAR-10 \cite{r1}, SVHN \cite{r2}, STL-10 \cite{r3}, and ImageNet \cite{r4}. MLAAN is integrated with DGL \cite{r6} and InfoPro \cite{r7} to assess performance and compare with BP \cite{r12} and original supervised local learning methods. Performance is compared under consistent $K$ conditions, typically $K=3$, ensuring rigorous and fair evaluation.

\subsection{Comparison with the SOTA results}

\subsubsection{Results on Image Classification Benchmarks.}We evaluate MLAAN's performance, with results in Table \ref{tab:table1}. On CIFAR-10 \cite{r1}, increasing local modules enhances MLAAN's benefits. For ResNet-32 (k=16) \cite{r5}, the Test Error decreases from 14.08 and 12.93 to 6.36 and 6.37, marking a 55\% and 51\% improvement over the BP \cite{r12} baseline. For ResNet-110 (k=55) \cite{r5}, MLAAN with DGL \cite{r6} and InfoPro \cite{r7} improves performance by 61\% and 56\%, respectively. On STL-10 \cite{r3}, MLAAN achieves over 30\% and 34\% improvements under the same situations, while on SVHN \cite{r2}, gains exceed 42\% and 49\%.
\begin{table}[h]
\centering
\begin{tabular}{cccc}
\toprule
Backbone                                                                     & Method         & Top1 Error                       & Top5 Error           \\ \midrule
\multirow{3}{*}{\begin{tabular}[c]{@{}c@{}}ResNet-34\\ (K=17)\end{tabular}}  & DGL            & 39.93                            & 15.58                \\
                                                                             & BP             & 25.38                            & 7.90                 \\
                                                                             & \textbf{MLAAN} & \textbf{25.31(↓0.07)} & \textbf{7.88(↓0.02)} \\ \midrule
\multirow{3}{*}{\begin{tabular}[c]{@{}c@{}}ResNet-101\\ (K=34)\end{tabular}} & DGL            & 35.58                            & 13.71                \\
                                                                             & BP             & 22.03                            & 5.93                 \\
                                                                             & \textbf{MLAAN} & \textbf{21.69(↓0.34)} & \textbf{5.72(↓0.21)} \\ \midrule
\multirow{3}{*}{\begin{tabular}[c]{@{}c@{}}ResNet-152\\ (K=51)\end{tabular}} & DGL            & 35.80                            & 14.15                \\
                                                                             & BP             & 21.60                            & 5.92                 \\
                                                                             & \textbf{MLAAN} & \textbf{21.34(↓0.26)} & \textbf{5.74(↓0.18)} \\ \bottomrule
\end{tabular}
\caption{Results on the validation set of ImageNet}
\label{tab:table2}
\end{table}
\begin{figure*}[!t]
\begin{minipage}[b]{0.24\linewidth}
  \centering
  \includegraphics[width=2.5cm]{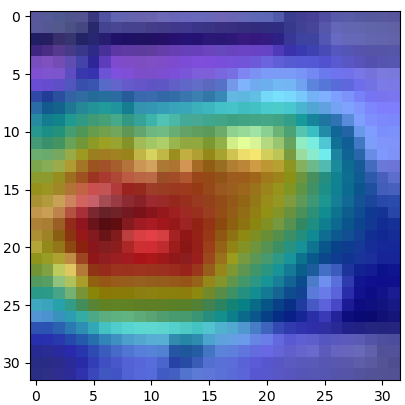}
  \centerline{(a)}\medskip
\end{minipage}
\begin{minipage}[b]{0.24\linewidth}
  \centering
  \includegraphics[width=2.5cm]{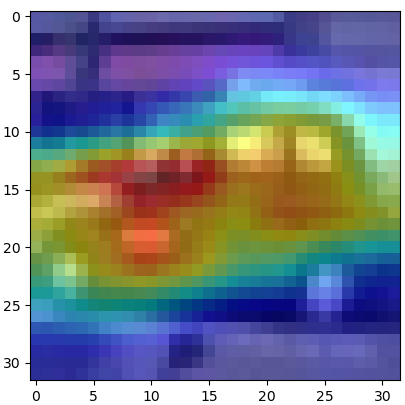}
  \centerline{(b)}\medskip
\end{minipage}
\begin{minipage}[b]{0.24\linewidth}
  \centering
  \includegraphics[width=2.5cm]{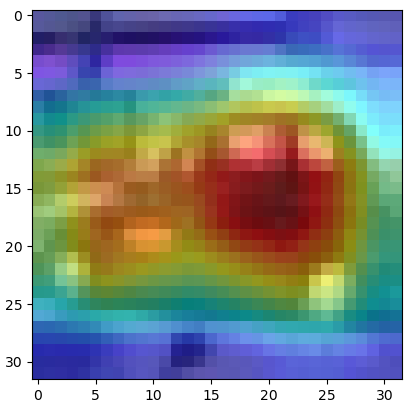}
  \centerline{(c)}\medskip
\end{minipage}
\begin{minipage}[b]{0.24\linewidth}
  \centering
  \includegraphics[width=2.5cm]{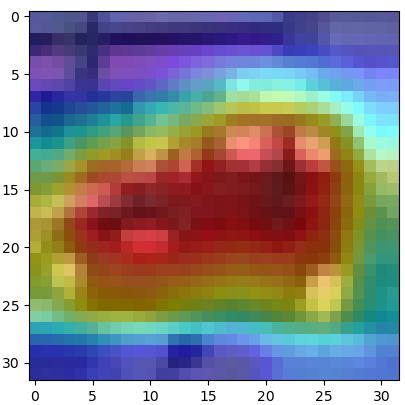}
  \centerline{(d)}\medskip
\end{minipage}
\caption{Visualization of feature maps. (a) Feature map of DGL. (b) Feature map of DGL with LAM. (c) Feature map of DGL with MLM. (d) Feature map of DGL with MLAAN.}
\label{fig:5}
\end{figure*}
Subsequently, we access MLAAN on ImageNet \cite{r4} using ResNet-34 \cite{r5} with 17 local modules. Incorporating MLAAN significantly improves performance, as shown in Table \ref{tab:table2}, reducing both Top1 and Top5 Error by 0.3\%, comparable to the BP \cite{r12} baseline. Further experiments' performance is also over BP \cite{r12}. ResNet-101 \cite{r5} sees Top1 and Top5 Error reductions of 1.5\% and 3.5\%, while ResNet-152 \cite{r5} shows decreases of 1.2\% and 3.0\%.

These results demonstrate MLAAN's ability to improve deep learning models and outperform BP even with more local modules, reinforcing its effectiveness in improving accuracy and robustness in complex and deep architectures.

\subsubsection{Memory Consumption.}We investigate the GPU memory usage of BP \cite{r12} and MLAAN on CIFAR-10 \cite{r1} and ImageNet \cite{r4}, with results in Table \ref{tab:table3}.

For ResNet-32 and ResNet-110 \cite{r5}, usage decreases by 22.3\% and 70.2\%, compared to BP \cite{r12}. ResNet-34, ResNet-101, and ResNet-152 \cite{r5} also show reductions of 15.4\%, 13.1\%, and 12.5\%, respectively. These findings demonstrate that MLAAN performs better with less GPU memory usage than BP \cite{r12}.
\begin{table}[h]
\centering
\setlength{\tabcolsep}{1.1mm}
\begin{tabular}{cccc}
\toprule
Dataset & Network                                                                      & Method         & GPU Memory(GB)          \\ \midrule
\multirow{4}{*}{CIFAR-10} & \multirow{2}{*}{\begin{tabular}[c]{@{}c@{}}ResNet-32\\ (K=16)\end{tabular}} & BP & 3.37G  \\
        &                                                                              & \textbf{MLAAN} & \textbf{2.62G(↓22.3\%)}  \\
        & \multirow{2}{*}{\begin{tabular}[c]{@{}c@{}}ResNet-110\\ (K=55)\end{tabular}} & BP             & 9.26G                   \\
        &                                                                              & \textbf{MLAAN} & \textbf{2.76G(↓70.2\%)}  \\ \midrule
\multirow{6}{*}{ImageNet} & \multirow{2}{*}{\begin{tabular}[c]{@{}c@{}}ResNet-34\\ (K=17)\end{tabular}} & BP & 10.74G \\
        &                                                                              & \textbf{MLAAN} & \textbf{9.09G(↓15.4\%)}  \\
        & \multirow{2}{*}{\begin{tabular}[c]{@{}c@{}}ResNet-101\\ (K=34)\end{tabular}} & BP             & 20.64G                  \\
        &                                                                              & \textbf{MLAAN} & \textbf{17.93G(↓13.1\%)}  \\
        & \multirow{2}{*}{\begin{tabular}[c]{@{}c@{}}ResNet-152\\ (K=51)\end{tabular}} & BP             & 26.29G                  \\
        &                                                                              & \textbf{MLAAN} & \textbf{23.01G(↓12.5\%)} \\ \bottomrule
\end{tabular}
\caption{Comparison of GPU memory usage between BP and MLAAN on CIFAR-10 and ImageNet.}
\label{tab:table3}
\end{table}

\subsection{Ablation Study}
\subsubsection{Comparison of Different Module Combinations.}We conduct experiments with ResNet-32 (K=16) \cite{r5} as the backbone, using DGL \cite{r6} as the baseline, to evaluate MLM, LAM, and MLAAN. The results are summarized in Table \ref{tab:table4}.

Without MLM and LAM, the Test Error is 14.08. Incorporating MLM individually reduces the Test Error by 5.62 and 6.44, respectively. Combining both modules results in a further reduction to 6.80, a substantial decrease of 7.28. These findings highlight the significant contributions of MLM and LAM, demonstrating MLAAN's remarkable performance through their integration.

\begin{table}[h]
\centering
\setlength{\tabcolsep}{8mm}
\begin{tabular}{ccc}
\toprule
MLM & LAM & Test Error  \\ \midrule
\XSolid                         & \XSolid                     & 14.08       \\
\Checkmark                          & \XSolid                      & 8.46(↓5.62) \\
\XSolid                          & \Checkmark                      & 7.64(↓6.44) \\
\Checkmark                          & \Checkmark                      & 6.36(↓7.28) \\ \bottomrule
\end{tabular}
\caption{Abalation study of our method on CIFAR-10.}
\label{tab:table4}
\end{table}

\subsubsection{Comparative Analysis of Unfair Epochs.}Undeniably, the utilization of MLAAN introduces additional Wall Time, prompting us to conduct a comparative analysis of unfair epochs. The findings can be observed in Table \ref{tab:table5}. When using DGL \cite{r6} as the baseline, after incorporating MLAAN, it only takes 300 epochs to reduce the Test Error by 45\% for a Backbone of ResNet-32 (K=16) \cite{r5} and 52\% for ResNet-110 (K=55) \cite{r5}.
\begin{figure*}[!t]
\begin{minipage}[b]{0.5\linewidth}
  \centering
  \includegraphics[width=8.2cm]{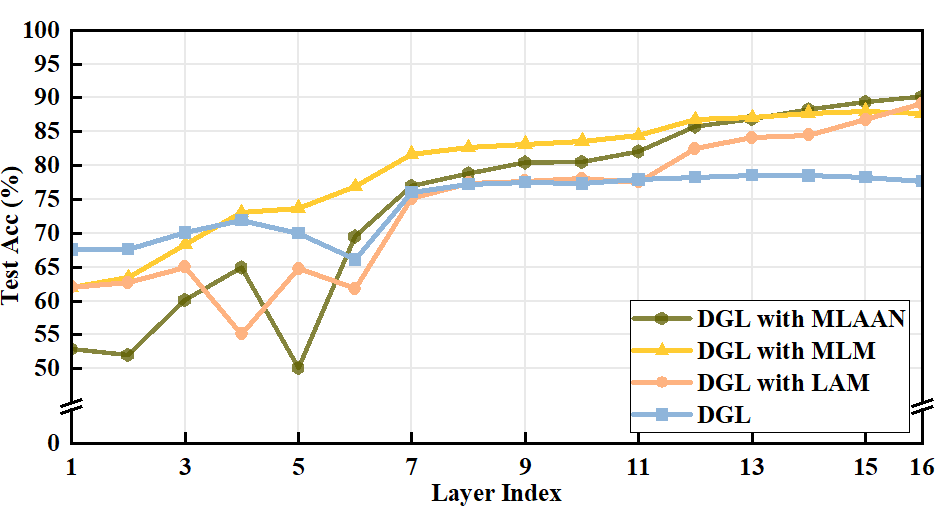}
  \centerline{(a)}\medskip
\end{minipage}
\begin{minipage}[b]{0.5\linewidth}
  \centering
  \includegraphics[width=8.19cm]{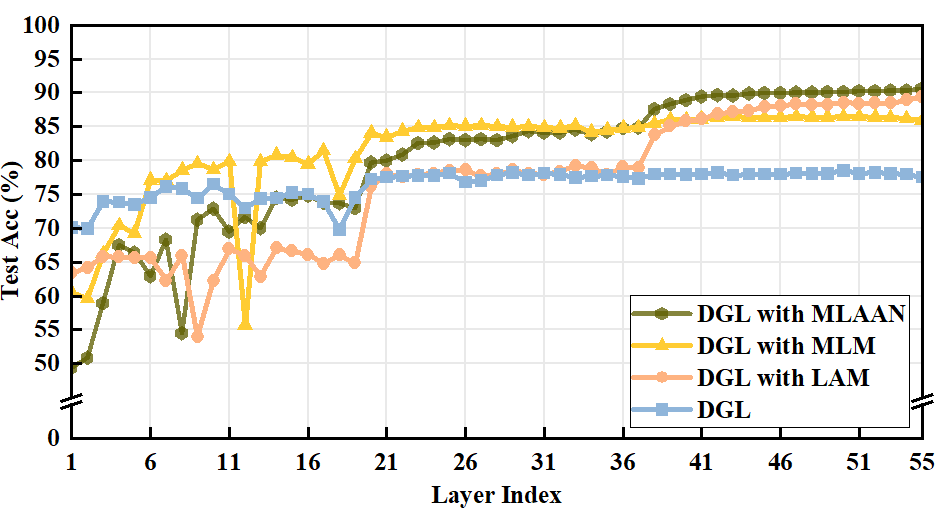}
  \centerline{(b)}\medskip
\end{minipage}
\caption{Comparison of layer-wise linear separability. (a) Linear Separability of RestNet-32 on CIFAR-10. (b) Linear Separability of RestNet-110 on CIFAR-10.}
\label{fig:6}
\end{figure*}
\begin{figure*}[!t]
\begin{minipage}[b]{0.5\linewidth}
  \centering
  \includegraphics[width=8.35cm]{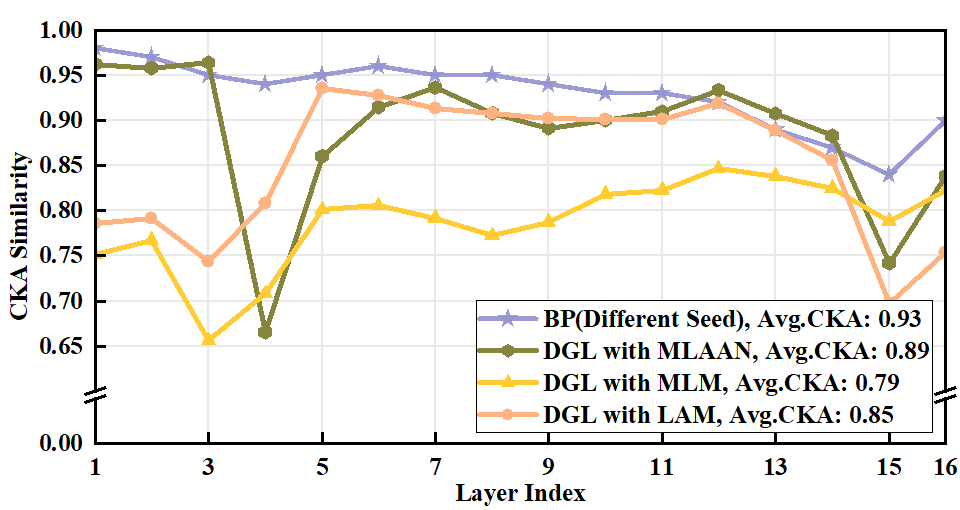}
  \centerline{(a)}\medskip
\end{minipage}
\begin{minipage}[b]{0.5\linewidth}
  \centering
  \includegraphics[width=8.35cm]{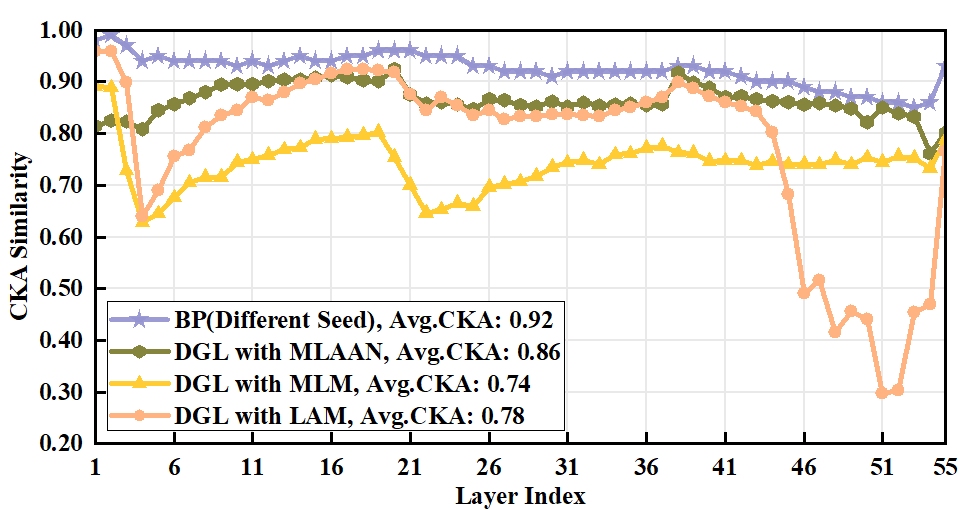}
  \centerline{(b)}\medskip
\end{minipage}
\caption{Comparison of layer-wise representation similarity. (a) Representation Similarity of RestNet-32 on CIFAR-10. (b) Representation Similarity of RestNet-110 on CIFAR-10.}
\label{fig:7}
\end{figure*}
\begin{table}[h]
\centering
\setlength{\tabcolsep}{0.7mm}
\begin{tabular}{cccc}
\toprule
\multirow{2}{*}{Dataset}  & \multirow{2}{*}{Method} & ResNet-32   & ResNet-110  \\
                          &                         & (K=16)      & (K=55)      \\ \midrule
\multirow{2}{*}{CIFAR-10} & DGL(Epochs=400)      & 14.08       & 14.45       \\
                          & DGL*(Epochs=300)     & 7.79(↓6.29) & 6.98(↓7.47) \\ \bottomrule
\end{tabular}
\caption{Comparison of Test Error with different epochs on the CIFAR-10. The * means the addition of MLAAN.}
\label{tab:table5}
\end{table}

\subsubsection{Comparison of Features in Different Methods.}To showcase the advanced capabilities of MLAAN, we conduct feature map analyses on different configurations, including DGL \cite{r6}, DGL with MLM, DGL with LAM, and DGL with MLAAN. The resulting figures can be found in Figure \ref{fig:5}. 

Upon analyzing them, we can observe that (a) is concentrated in specific regions, indicating the presence of significant information within those areas. Conversely, after the fusion of (b) and (c), (d) captures more comprehensive global features, including localized edge features. It follows that MLAAN can compensate for the shortcomings of other methods.

\subsubsection{Decoupled Layer Accuracy Analysis.}To further analyze MLAAN's impact, we freeze the main network parameters and train a linear classifier for each gradient-isolated layer. Results are shown in Figure \ref{fig:6} using DGL \cite{r6} as the baseline.

High classification accuracy in early layers can hinder global network performance by optimizing local objectives at the expense of features needed by later layers. MLAAN addresses this by reducing early layers' accuracy to capture more globally useful features, enhancing subsequent layers' generalization. 

\subsubsection{Representation Similarity Analysis.} We conduct Centered Kernel Alignment (CKA) \cite{r11} experiments. We incorporate MLAAN and calculate the CKA \cite{r11} similarity for each layer compared to BP \cite{r12}, then compute the mean values. Results are shown in Figure \ref{fig:7}.

MLAAN significantly enhances DGL's CKA similarity, particularly in early layers. The original method's poor performance arises from early layers focusing on local optimization, which reduces overall performance.

\section{Conclusion}

This paper introduces the Multilaminar Leap Augmented Auxiliary Network (MLAAN), a novel approach to resolving issues related to the lack of global features and myopic perspectives that lead to poor performance in existing supervised local learning. MLAAN consists of Multilaminar Local Modules (MLM) and Leap Augmented Modules (LAM). MLM addresses the issue of insufficient global features in existing work. We further propose that LAM enhance the auxiliary network for inter-module information exchange, effectively utilizing deeper information to mitigate shortsightedness. We integrate our MLAAN into two SOTA supervised local learning methods and evaluate the performance of various deep network architectures across four widely used datasets. The results demonstrate that our approach significantly enhances the performance of original supervised local learning methods, even surpassing that of E2E training.

\section{Acknowledgments}
This work was supported by the National Natural Science Foundation of China
(52278154) and the Natural Science Foundation of Jiangsu (BK20231429).
\bigskip

\bibliography{aaai25}
\appendix
\section{Supplementary Material}
\subsection{Implement Detail}

Our experiments use ResNet-32 and ResNet-110 \cite{r5} as backbone networks. We employ the SGD optimizer \cite{r8} with Nesterov momentum \cite{r9} at 0.8 and an L2 weight decay of 1e-4. ResNet-32 \cite{r5} is divided into 16 modules, while ResNet-110 \cite{r5} is segmented into 55 modules, each with its own parameters and auxiliary network.

Different hyperparameters are used for each dataset: batch size is 1024 for CIFAR-10 \cite{r1} and SVHN \cite{r2}, and 128 for STL-10 \cite{r3}. Training lasts 400 epochs with initial learning rates of 0.8 for CIFAR-10 \cite{r1} and SVHN \cite{r2}, and 0.1 for STL-10 \cite{r3}, following a cosine annealing scheduler \cite{r10}. For ImageNet \cite{r4}, we use ResNet-34, ResNet-101, and ResNet-152 \cite{r5}, training for 150 epochs with an initial learning rate of 0.05 and a batch size of 128. We maintain consistency with other training configurations as previously described for CIFAR-10 \cite{r1}.



\subsection{Generalization Study}

In this section, we aim to investigate the generalization performance of our proposed MLAAN method. To evaluate its effectiveness, we utilize the checkpoints trained on the CIFAR-10 \cite{r1} and test them on the STL-10 \cite{r3}, taking inspiration from previous work \cite{r34}.

As shown in Table \ref{tab:table6}, we observe a notable disparity in the generalization abilities of DGL \cite{r6} and BP \cite{r12}. However, with the incorporation of our MLAAN method, we witness a significant improvement in test accuracy, surpassing even that of BP \cite{r12}. These results indicate that MLAAN, by facilitating information interaction among local modules, effectively enhances the generalization ability of supervised local learning methods.

\begin{table}[h]
\centering
\setlength{\tabcolsep}{2.8mm}
\begin{tabular}{ccc}
\toprule
Method        & ResNet-32(K=16) & ResNet-110(K=55) \\ \midrule
BP            & 35.98           & 36.78            \\
DGL           & 31.95           & 33.16            \\
\textbf{DGL*} & \textbf{37.65}       & \textbf{40.26}        \\ \bottomrule
\end{tabular}
\caption{Generalization study. Checkpoints are trained on the CIFAR-10 dataset and tested on the STL-10 dataset. The data in the table represents the test accuracy.}
\label{tab:table6}
\end{table}

These findings emphasize the efficacy of our MLAAN method in improving the generalization capabilities of supervised local learning, ultimately leading to enhanced overall performance in the image classification task.
\end{document}